\pdfoutput=1

\documentclass[11pt]{article}
\usepackage[dvipsnames]{xcolor}

\usepackage{EMNLP2022}

\usepackage{times}
\usepackage{latexsym}

\usepackage[T1]{fontenc}

\usepackage[utf8]{inputenc}

\usepackage{microtype}

\usepackage{inconsolata}

\usepackage{hyperref}
\usepackage{url}

\usepackage{amssymb}
\usepackage{amsbsy}
\usepackage{amsmath}
\usepackage{amsthm}
\usepackage{amsfonts}
\usepackage{latexsym}
\usepackage{subcaption}
\usepackage{wrapfig}
\usepackage{multirow}

\usepackage{graphicx}
\usepackage{tabularx}
\usepackage{booktabs}
\usepackage{ulem}
\usepackage{caption}
\usepackage{algorithm}
\usepackage{algorithmicx, algpseudocode}
\usepackage{mathtools}
\usepackage{arydshln}
\usepackage{titlesec}

\usepackage{amsmath,amsfonts,bm}

\def\eqref#1{equation~\ref{#1}}

\def\1{\bm{1}}

\def\rvx{{\mathbf{x}}}
\def\rvy{{\mathbf{y}}}

\DeclareMathAlphabet{\mathsfit}{\encodingdefault}{\sfdefault}{m}{sl}
\SetMathAlphabet{\mathsfit}{bold}{\encodingdefault}{\sfdefault}{bx}{n}

\def\gL{{\mathcal{L}}}

\def\gX{{\mathcal{X}}}
\def\gY{{\mathcal{Y}}}

\newcommand{\E}{\mathbb{E}}

\newcommand{\R}{\mathbb{R}}

\newcommand{\prx}{r_x(\rvx)}

\newcommand{\pry}{r_y(\rvy)}

\newcommand{\pryp}{r_y(\rvy^{\prime})}
\newcommand{\hy}{\hat{\rvy}}

\newcommand{\std}[1]{{\scriptsize $\pm${#1}}}
\newcommand{\up}[1]{\textcolor{OliveGreen}{\small \ $\uparrow${#1}}}
\newcommand{\down}[1]{\textcolor{Maroon}{\small \ $\downarrow${#1}}}

\newcommand\eqfootnote[1]{%
  \begingroup
  \renewcommand\thefootnote{}\footnote{#1}%
  \addtocounter{footnote}{-1}%
  \endgroup
}

\title{Prompt Consistency for Zero-Shot Task Generalization}

\author{Chunting Zhou$^{*1}$, Junxian He$^{*1}$, Xuezhe Ma$^{2}$, Taylor Berg-Kirkpatrick$^{3}$, Graham Neubig$^{1}$ \\
  $^{1}$Language Technologies Institute, Carnegie Mellon University \\
    $^{2}$ Information Sciences Institute, University of Southern California \\
  $^{3}$Department of Computer Science and Engineering, University of California San Diego\\
  \texttt{\{chuntinz,junxianh,gneubig\}@cs.cmu.edu} \\
  \texttt{xuezhema@isi.edu, tberg@eng.ucsd.edu}\\}

\begin{document}
\maketitle
\begin{abstract}
One of the most impressive results of recent NLP history is the ability of pre-trained language models to solve new tasks in a \emph{zero-shot} setting. To achieve this, NLP tasks are framed as natural language prompts, generating a response indicating the predicted output. Nonetheless, the performance in such settings often lags far behind its supervised counterpart, suggesting a large space for potential improvement. In this paper, we explore methods to utilize unlabeled data to improve zero-shot performance. Specifically, we take advantage of the fact that multiple prompts can be used to specify a single task, and propose to regularize \emph{prompt consistency}, encouraging consistent predictions over this diverse set of prompts. Our method makes it possible to fine-tune the model either with extra unlabeled training data, or directly on test input at inference time in an unsupervised manner. In experiments, our approach outperforms the state-of-the-art zero-shot learner, T0~\citep{sanh2021multitask}, on 9 out of 11 datasets across 4 NLP tasks by up to 10.6 absolute points in terms of accuracy. The gains are often attained with a small number of unlabeled examples.\eqfootnote{\hspace{-1.5mm}$^{*}$Equal contribution. Order determined by swapping the one in~\citet{he2021towards}.}\footnote{Code is available at \href{https://github.com/violet-zct/swarm-distillation-zero-shot}{https://github.com/violet-zct/swarm-distillation-zero-shot}.}
\end{abstract}

\section{Introduction}
\label{sec:intro}
While the past decade has demonstrated that pretrained language models (PLMs) are powerful tools for improving generalization from training datasets to test datasets~\citep{devlin-etal-2019-bert,liu2019roberta,raffel2020exploring}, more recent work has shown that they can even perform \emph{zero-shot generalization to new tasks} without any annotated examples ~\citep{brown2020language,wei2021finetuned,sanh2021multitask}. 
These systems leverage natural language prompts that specify the task for the model and represent different tasks in a unified format~\citep{liu2021pre}.
Zero-shot task generalization suggests a path towards generic systems that perform a wide variety of NLP tasks with no annotated examples.
However, while enticing conceptually, zero-shot performance often remains relatively low compared to systems trained using even a small amount of task-specific labeled data.


In this paper, we examine methods to make PLMs better zero-shot learners using unlabeled text. Our work is motivated by consistency training methods that regularize model predictions to be invariant to perturbation (e.g. noise or paraphrasing) of the input examples. Consistency training is widely used in semi-supervised learning literature as an effective technique to utilize unannotated examples~\citep{bachman2014learning,sajjadi2016regularization,zhai2019s4l,xie2020unsupervised}. 
It is often understood as a type of smoothness regularization or data augmentation~\citep{xie2020unsupervised} and attains strong performance in semi-supervised learning.
Instead of example-level consistency, we propose to regularize \emph{prompt consistency}, where a model is regularized to make the same prediction across a diverse set of synonymous task prompts. 
Prompt consistency regularization makes sense intuitively since PLMs should be robust across synonymous prompts, whereas it is known that model predictions are empirically very sensitive to the wording of the task prompts~\citep{jiang-etal-2020-know}.

\begin{figure*}[!t]
    \centering
    \includegraphics[width=1\textwidth]{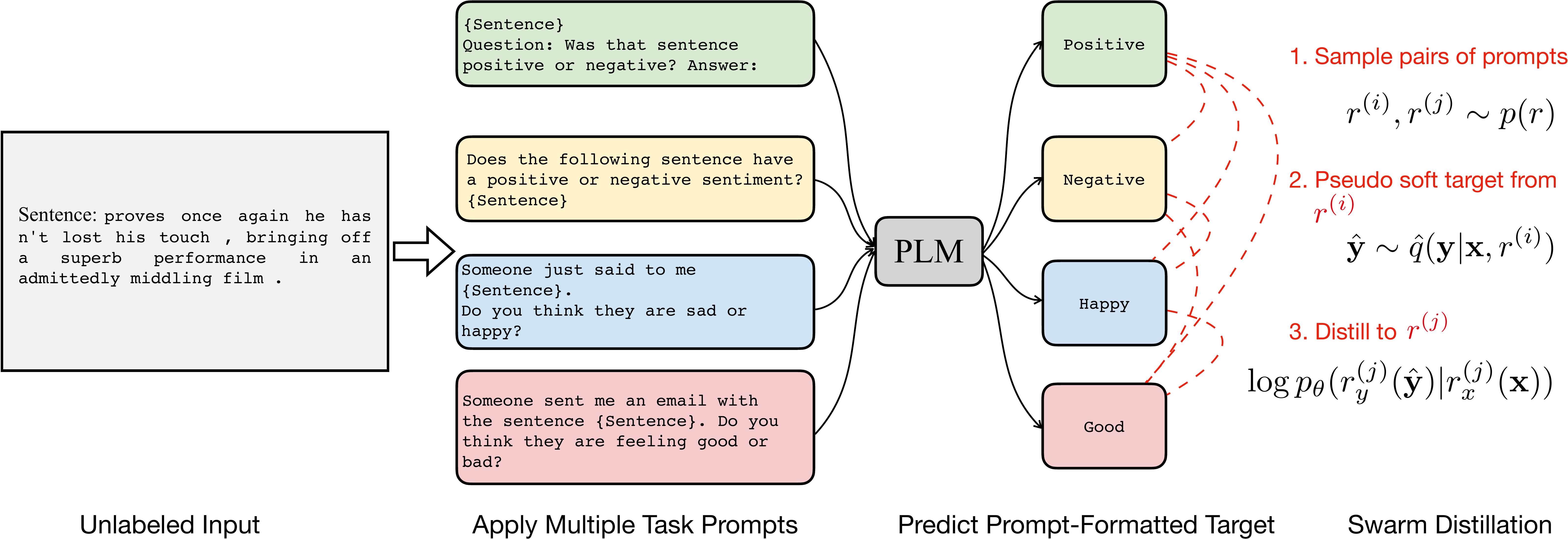}
    \caption{An example of the proposed approach in an sentiment classification task. We apply multiple synonymous prompts to the unlabeled example, then we regularize the consistency of the predictions from different prompts, through our swarm distillation loss 
    as detailed in Eq.~\ref{eq:pairwise-distill}.
    }
    \label{fig:intro-model}
\end{figure*}

%
Specifically, we design a pairwise distillation loss that encourages consistency between every pair of prompts (Figure~\ref{fig:intro-model}). We refer to our method as \emph{swarm distillation}, and it has the advantage of being fully unsupervised, only requiring unannotated inputs.
Notably, unannotated examples are often relatively easy to collect.
Drafting several prompts for a task is also far cheaper than annotating labels for each example -- in fact, there are already well-designed prompts available for a wide range of NLP tasks~\citep{bach2022promptsource}.

Previous work on example-level consistency regularization typically minimizes a consistency loss along with a supervised loss in a semi-supervised setting~\citep{miyato2018virtual,xie2020unsupervised}.
Recently,~\citet{elazar-etal-2021-measuring} performed experiments optimizing a prompt consistency loss in the context of a relation prediction task, also incorporating a supervised version of the masked language model pretraining objective.
In contrast, we (1) optimize a novel prompt consistency loss alone, making our approach completely unsupervised and agnostic to the model's pretraining objective, and (2) experiment on and demonstrate the practicality of such an approach for a broad variety of NLP tasks.
Notably, this unsupervised setting poses additional learning challenges: without explicit supervision, the model may suffer from catastrophic forgetting and even exhibit a form of collapse where the model always makes the same predictions for any input. To address this issue, we adopt two simple strategies: (1) we utilize parameter-efficient tuning techniques~\citep{houlsby2019parameter,he2021towards} to only update a small number of extra parameters, naturally mitigating catastrophic forgetting by fixing the original PLM parameters; (2) we propose an unsupervised criterion to select the model checkpoint before it falls into a collapsed local optimum.  

In experiments, we build our method on top of a state-of-the-art zero-shot task learner, T0~\citep{sanh2021multitask}, and validate its performance on 11 datasets from 4 NLP tasks: natural language inference, coreference resolution, word sense disambiguation, and sentence completion. We perform experiments under two secenarios: (1) training the model with unlabeled training data; or (2) tuning the model with unlabeled test inputs directly. In both settings,
we show that our swarm distillation method improves the accuracy of the 3B-parameter T0 model on 9 out of 11 datasets by up to 10.6 absolute points. We further scale model size up to 11B parameters, and demonstrate that our approach outperforms the 11B-parameter T0 model on 4 out of 4 datasets. Remarkably, analysis implies that these gains are often possible with only tens of examples, suggesting a small computation overhead.

\section{Prompt-based Zero-Shot Task Generalization}
\label{sec:overview}
Given a task where the input is denoted as $\rvx \in \gX$ and the goal is to predict $\rvy \in \gY$, 
we focus on the zero-shot task generalization setting: we aim to feed a PLM with $\rvx$ to predict $\rvy$, where the PLM is never trained on the specific task to be performed. Zero-shot task generalization goes beyond traditional dataset generalization, as the model must generalize to new functions $f:\gX \rightarrow \gY$ as opposed to new input examples, $\rvx$. Recently, the development of prompting methods has advanced zero-shot task generalization by representing different tasks in a unified format~\citep{liu2021pre}, and several prompt-based approaches have attained reasonable zero-shot performance~\citep{brown2020language,sanh2021multitask,wei2021finetuned}. 

A prompt $r$ consists of an input template $r_x$, an output template $r_y$, and metadata to re-format the original $\rvx$ and $\rvy$ into new prompt-formatted input and target, $\prx$ and $\pry$. 
For example, as shown in Figure~\ref{fig:intro-model}, in a sentiment classification task where we must predict positive or negative sentiment of the text, the input includes the field \texttt{Sentence} and the target consists of the field \texttt{Label}. An input template could be ``Does the following sentence have a positive or negative sentiment? \texttt{\{Sentence\}}'', and the target template is ``\texttt{Choices[\{label\}]}''.
Here \texttt{Choices} is the metadata that is a list containing [\texttt{Positive}, $\texttt{Negative}$] to correspond to the numeric label ids. We note that such metadata is prompt-specific and can differ with different prompts for the same task -- for instance, in Figure~\ref{fig:intro-model} 
the \texttt{Choices} list of the last prompt on the bottom is [\texttt{Good}, $\texttt{Bad}$]. In prompt-based approaches the PLM 
models the conditional probability $q(\rvy|\rvx, r)$ through $p_{\theta}(\pry | \prx)$ where $\theta$ denotes the model parameters. 
In classification tasks where $\gY$ is a finite label set, $q(\rvy|\rvx, r)$ is normalized over the possible labels at inference time to predict $\rvy$:
\begin{equation}
\label{eq:decode}
q(\rvy | \rvx, r) = \frac{p_{\theta}(\pry | \prx)}{\sum_{\rvy^{\prime}\in\gY} p_{\theta}(\pryp | \prx)}.
\end{equation}
In generation tasks where $\gY$ is an infinite sequence space, the target template is typically instantiated as the target itself, i.e. $p_{\theta}(\pry | \prx) = p_{\theta}(\rvy | \prx)$, then the output can be directly decoded through sequence decoding approaches.
Through designing such prompts for each task, all NLP tasks share the same data format, and models trained on one task may generalize to others.

\section{Prompt Consistency Training}
\subsection{Problem Definition}
In this paper, we aim to explore unannotated examples to improve prompt-based zero-shot task generalization. Formally, we are given an unlabeled dataset in the task of interest $\{\rvx_1, \rvx_2, \cdots, \rvx_N\}$, and we assume the dataset has $K$ different prompts, $\{(r_x^{(1)}, r_y^{(1)}), \cdots, (r_x^{(K)}, r_y^{(K)})\}$. 
Our goal is to utilize these resources and adapt a PLM to predict $\pry$ conditioned on $\prx$. Unlabeled inputs are often available in practice, we consider two such scenarios in the paper.

First, we consider the case when unannotated examples from a non-test set are available. For many NLP tasks their inputs are plain text such as reviews, documents, or questions and can be easily collected (less so for other NLP tasks, like natural language inference the inputs are paired hypotheses and premises that can be non-trivial to obtain automatically). In this paper, we test this setting by utilizing the inputs of the training dataset. This is similar to~\citet{schick-schutze-2021-exploiting,schick-schutze-2021-just} where they directly use the inputs of the training split as unlabeled resources to help few-shot learning.  

Second is the case when unannotated test inputs are available. This is almost always true for any task. We use the test split to mimic the setting. While the limited number of unlabeled examples could potentially limit the effectiveness of some unsupervised learning methods, we show in \textsection\ref{sec:analysis} that our method is effective even with tens to hundreds of unlabeled examples.

On the other hand, a diverse set of prompts is not exceedingly difficult to collect practically -- drafting prompts for each task is easier than annotating labels for many examples.
In fact, the community efforts have pushed out a Public Pool of Prompts (P3)
that contains thousands of prompts for hundreds of NLP datasets already~\citep{bach2022promptsource}.

\subsection{The Prompt Consistency Loss}
Consistency regularization is a method that creates different views (e.g.~paraphrases of text) of the input and regularizes the outputs to be close to each other, and has achieved significant success in semi-supervised learning~\citep{clark-etal-2018-semi,xie2020unsupervised,xie2020self}. 
While previous methods use an additional module to perturb each example and then optimize example-level consistency, we propose to optimize prompt-level consistency which (1) is conceptually simple, and (2) can mitigate the fact that the predictions of PLMs are typically inconsistent with different prompts for the same task~\citep{jiang-etal-2020-know,elazar-etal-2021-measuring}. Intuitively, we propose to regularize the predictions of different prompts for a given input to be close to each other, using a pairwise distillation loss to draw the predictions from one prompt closer to those from the other. 
Concretely, we randomly sample a few pairs of prompts and distill the pseudo target $\hy$ from one prompt $r^{(i)}$ to the other prompt $r^{(j)}$,
as illustrated in Figure~\ref{fig:intro-model}. The loss function is defined as:
\begin{equation}
\label{eq:pairwise-distill}
\begin{split}
\gL = -&\E_{\rvx\sim p_{d}(\rvx)}\E_{r^{(i)}, r^{(j)}\sim p(r)} \\
&\E_{\hy\sim \hat{q}(\rvy|\rvx, r^{(i)})}\log p_{\theta}(r_y^{(j)}(\hy) | r_x^{(j)}(\rvx)),
\end{split}
\end{equation}
where $p_d(\rvx)$ is the empirical data distribution, $p(r)$ is a uniform distribution over possible prompts, and $\hat{q}(\rvy|\rvx, r)$ is the conditional target distribution defined as in Eq.~\ref{eq:decode} but with a stopping gradient operator. We do not propagate gradients to $\hat{q}(\rvy|\rvx, r^{(i)})$ following \citet{miyato2018virtual} and \citet{xie2020unsupervised}.\footnote{Note that $\hat{q}(\rvy|\rvx, r)$ still changes as we train the model.} 
Stopping the gradient of one side in a pairwise consistency loss is also shown to help mitigate the collapse issue where all inputs lead to the same predictions~\citep{chen2021exploring}. 
Different from traditional distillation that distills from a teacher model to a student model~\citep{hinton2015distilling}, or previous consistency training that a single teacher distills to several students~\citep{clark-etal-2018-semi,xie2020unsupervised}, we perform distillation among a swarm of prompts where each prompt is a teacher and student at the same time, thus we term our method as \emph{swarm distillation}. In our implementation, we approximate the expectation over the paired prompts $(r^{(i)}, r^{(j)})$ with $k$ randomly sampled pairs 
for training efficiency.  

Prompt consistency is related to example-level consistency when viewing different prompt-formatted inputs $r_x^{(i)}(\rvx)$ as separated views of the same example, thus our swarm distillation approach shares spirit with previous work on example-level consistency training and can be understood similarly from the perspective of unsupervised data augmentation, smoothness regularization, or label propagation~\citep{xie2020unsupervised}. In this paper, we focus on classification tasks where $\gY$ is a finite label set, while Eq.~\ref{eq:pairwise-distill} can be directly applied to sequence generation tasks as well with sequence distillation~\citep{kim-rush-2016-sequence}. 

Our approach differs from previous consistency training methods which often combine an unsupervised consistency loss with a supervised loss in a semi-supervised setting~\citep{miyato2018virtual,clark-etal-2018-semi,xie2020unsupervised}.~\citet{elazar-etal-2021-measuring} try to improve prompt consistency for a relation filling task with a pairwise two-sided KL divergence loss, while they also optimize a supervised version of the original PLM objective that turns out to be important. In contrast, our approach minimizes the swarm distillation loss in Eq.~\ref{eq:pairwise-distill} alone, and therefore is completely unsupervised and agnostic to the pretraining objective. However, this setting also poses challenges in learning, which we discuss next.   

\subsection{Training}
\label{sec:train}
Being trained without explicit supervision, the PLM may forget what it learns during pretraining since the unsupervised consistency loss is different from the pretraining objective. Also, we note that prompt consistency may be achieved with a trivial solution -- if the predictions from each example and each prompt collapse to the same label then maximal consistency among prompts can be reached. To mitigate such catastrophic forgetting and collapse issues, we propose two techniques:

\begin{figure}[!t]
    \centering
    \includegraphics[width=0.3\textwidth]{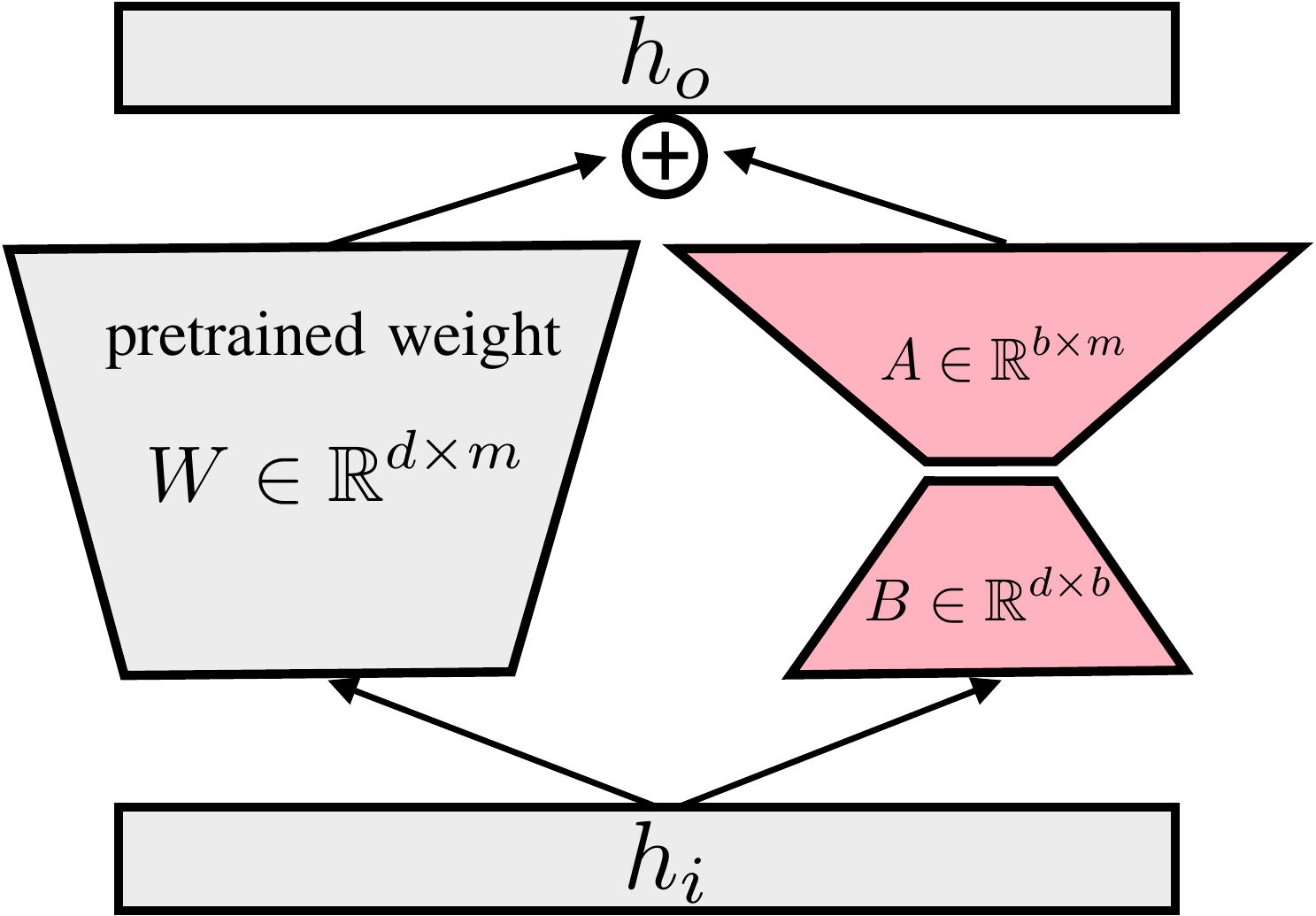}
    \caption{A diagram of LoRA in the FFN sublayer. Only the LoRA parameters, $A$ and $B$, are updated during training. 
    }
    \label{fig:lora}
    \vspace{-12pt}
\end{figure}

\paragraph{Parameter-efficient tuning:}
It has recently been observed that updating a small number of added parameters in a PLM is able to achieve comparable performance to tuning all the parameters~\citep{houlsby2019parameter,li-liang-2021-prefix,hu2021lora,he2021towards}. Parameter-efficient tuning methods naturally mitigate catastrophic forgetting and collapse through fixing the original PLM parameters. 
Specifically, we use LoRA~\citep{hu2021lora}, a low-rank adaptation method for PLMs. As shown in Figure~\ref{fig:lora}, LoRA learns a low-rank approximation of the pretrained matrix updates: given a pretrained weight matrix $W\in\R^{d\times m}$, LoRA learns to update it as $W \leftarrow W + \alpha BA$, where $B \in \R^{d\times b}, A\in\R^{b\times m}$ are low-rank matrices and $\alpha$ is a hyperparameter, and only $B$ and $A$ are updated during training. $b\ll d$ is referred to as the \emph{bottleneck dimension}. Following~\citet{he2021towards}, we apply LoRA to the feed-forward weight matrices of every layer in the pretrained transformer~\citep{vaswani2017attention} model. 
In our preliminary experiments, we found that LoRA is less likely to suffer from collapse, while on some datasets the model still collapses in the end even though it learns well in the middle. This motivates us to develop a criterion to select the model checkpoint 
before it falls into a collapsed local optimum, 
which we describe next.

\paragraph{Unsupervised model selection criterion:}
\label{sec:dev:metric}
Our zero-shot setting does not have labeled validation data for model selection, and the swarm distillation objective is not an ideal selection criterion since it is minimized at collapse.  
Therefore, we would like to have an unsupervised criterion that encourages consistency but simultaneously penalizes collapse. 
With that in mind, we focus on Fleiss' kappa~\citep{fleiss1971measuring}, a commonly used metric to assess the reliability of agreement. 
In our setting, Fleiss' kappa expresses the extent to which the amount of agreement among prompts exceeds what would be expected if all prompts made their predictions according to the marginalized distribution of labels. This design computes a notion of ``relative consistency'' and naturally penalizes collapse.
Formally, let $n_{ij}$ be the number of prompts that predict the $j$-th label for the $i$-th example. There are a total of $NK$ predictions where $N$ is the number of examples and $K$ is the number of prompts. Given an example $\rvx_i$, the agreement probability $p_i$ computes the normalized number of agreeing prompt pairs: 
\begin{equation}
p_i = \sum\nolimits_jn_{ij}(n_{ij}-1) / (K(K-1)),
\end{equation}
then the ``absolute consistency'' $\bar{P}$ is:
\begin{equation}
\label{eq:absolute}
\bar{P} = \sum\nolimits_{i=1}^Np_i / N.
\end{equation}
$\bar{P}$ is maximized in the case of collapse. However, Fleiss' kappa considers the marginalized distribution of labels: how likely are two prompts consistent if they make predictions randomly according to the marginalized label distribution? This chance probability $\bar{P}_e$ is:
\begin{equation}
\bar{P}_e=\sum\nolimits_jq^2_j, \qquad q_j=\sum\nolimits_{i=1}^Nn_{ij}/(NK),
\end{equation}
where $q_j$ represents the marginalized distribution of labels, i.e. $p(\rvy=j)$. $\bar{P}_e$ is large when collapse happens and one label dominates in the entire corpus. Finally, Fleiss' kappa is computed as:
\begin{equation}
\label{eq:kappa}
\kappa = \frac{\bar{P}-\bar{P}_e}{1-\bar{P}_e},
\end{equation}
where $1-\bar{P}_e$ gives the degree of consistency that is attainable above chance, $\bar{P}-\bar{P}_e$ gives the degree of consistency actually achieved above chance. $\kappa$ ranges from -1 to 1. Eq.~\ref{eq:kappa} naturally penalizes collapse, and in our experiments, we always observe a monotonic decrease of $\kappa$ when collapse happens. Therefore, we select the model checkpoint after which $\kappa$ monotonically decreases.\footnote{In most of the settings, this criterion is equivalent to using maximal $\kappa$ as the criterion, except for few cases where the beginning of training exhibits large fluctuations in $\kappa$.} 
We emphasize that we perform validation on the data that the model is trained on and do not require an additional development dataset. 
We include ablation analysis for both LoRA and the model selection components in Appendix~\ref{app:lora-full} that shows that they are important for the success of our method.
\begin{table*}[!t]
    \centering
    \resizebox{1. \textwidth}{!}{
    \begin{tabular}{llcccccccc}
    \toprule
                                             &             & \multicolumn{2}{c}{T0-3B} & \multicolumn{2}{c}{Self Dist. (train)}                          & \multicolumn{2}{c}{Swarm Dist. (train)} & \multicolumn{2}{l}{Swarm Dist. (test)} \\
    Task                                     & Dataset     & \multicolumn{1}{c}{Ens.} & \multicolumn{1}{c}{Med.} & Ens.                   & Med.      & Ens.                   & Med.            & Ens.                  & Med.                  \\
    \midrule
    \multicolumn{1}{l}{\multirow{5}{*}{NLI}} & RTE         & 64.6                     & 64.1          & 64.9\std{0.2} & 63.8\std{0.1}            & 75.2\std{0.8}\up{10.6}              & 73.9\std{0.8}\up{9.8}             & 75.2\std{0.2}\up{10.6}             & 73.5\std{0.1}\up{9.4}             \\
    \multicolumn{1}{l}{}                     & CB          &  46.4 &  50.0         & 47.0\std{1.0} & 49.4\std{2.7}  &    47.6\std{1.0}\up{1.2}         & 48.2\std{0.0}\down{1.8}                       &            46.4\std{0.0}\up{0.0}           &    48.8\std{1.0}\down{1.2}                   \\
    \multicolumn{1}{l}{}                     & ANLI R1     & 34.6                     & 33.7           & 36.1\std{0.1} & 34.7\std{0.1}          & 38.4\std{0.5}\up{3.8}              & 35.7\std{0.4}\up{2.0}             & 38.5\std{0.3}\up{3.9}             & 35.7\std{0.5}\up{2.0}             \\
    \multicolumn{1}{l}{}                     & ANLI R2     & 33.7                     & 33.4          & 35.3\std{0.1} & 33.2\std{0.2}           & 37.9\std{0.8}\up{4.2}              & 36.6\std{0.5}\up{3.2}             & 37.7\std{0.2}\up{4.0}            & 35.4\std{0.4}\up{2.0}             \\
    \multicolumn{1}{l}{}                     & ANLI R3     & 34.7                     & 33.3          & 33.1\std{0.0} & 33.8\std{0.2}            & 34.0\std{0.3}\down{0.7}              & 34.6\std{0.1}\up{1.3}             & 34.1\std{0.2}\down{0.6}             & 33.5\std{0.0}\up{0.2}             \\
    \midrule
    \multirow{3}{*}{Compl.}                  & COPA        & 78.0                       & 79.0        & 82.3\std{0.6} & 78.2\std{0.3}                & 82.7\std{0.6}\up{4.7}              & 79.0\std{0.5}\up{0.0}              & 83.0\std{1.0}\up{5.0}                 & 79.7\std{0.6}\up{0.7}             \\
                                             & HellaSwag   & 27.8                     & 27.5          & 32.5\std{0.2} & 32.7\std{0.3}            & 34.2\std{0.2}\up{6.4}             & 33.4\std{0.2}\up{5.9}             & 33.7\std{0.6}\up{5.9}             & 33.2\std{0.3}\up{5.7}             \\
                                             & Story Cloze & 86.5                     & 85.1          & 89.6\std{0.0} & 88.7\std{0.0}            &       --                 &       --                & 87.3\std{0.1}\up{0.8}            & 86.9\std{0.2}\up{1.8}             \\
    \midrule
    \multirow{2}{*}{Coref.}                  & Wino.       & 50.9                     & 50.5           & 51.1\std{0.1} & 50.7\std{0.1}           & 52.0\std{0.3}\up{1.1}              & 51.4\std{0.0}\up{0.9}             & 52.1\std{0.3}\up{1.2}             & 51.2\std{0.2}\up{0.7}             \\
                                             & WSC         & 69.2                     & 64.4           & 69.2\std{0.0} & 64.6\std{0.3}            & 58.3\std{1.1}\down{10.9}              & 59.3\std{2.0}\down{5.1}             & 57.7\std{0.0}\down{11.5}             & 58.8\std{0.6}\down{5.6}             \\
    \midrule
    WSD                                      & WIC         & 50.3                     & 50.4          & 50.3\std{0.0} & 50.3\std{0.0}            & 55.4\std{1.1}\up{5.1}              & 54.4\std{0.7}\up{4.0}             & 55.5\std{0.8}\up{5.2}             & 54.8\std{0.5}\up{4.4}            \\
    \bottomrule 
    \end{tabular}}
    \caption{\label{tab:res-main}Accuracy results on the validation set of 11 NLP datasets based on the T0-3B model. Swarm Distillation (train) and Swarm Distillation (test) use the unlabeled training split and validation split of datasets to train the model respectively, corresponding to training-time and test-time tuning. The Story Cloze dataset does not have a training split and its self distillation results are from tuning on the validation split. We report the mean and std across 3 random runs, and also denote the absolute accuracy change compared to the T0-3B baseline.}
\vspace{-5pt}
\end{table*}
\section{Experiments}
Our experiments below are designed to (1) measure whether swarm distillation is able to improve zero-shot task generalization; 
and (2) analyze how much resource (number of prompts and unlabeled examples) our method demands. 
\subsection{General  Setup}
\label{sec:setup}
\paragraph{Datasets:}
Following~\citet{sanh2021multitask}, we evaluate our method on 11 NLP datasets across 4 unseen tasks. They are (1) natural language inference: ANLI~\citep{nie-etal-2020-adversarial} (there are three versions of ANLI with different levels of difficulty, which we denote as ANLI R1/R2/R3), CB~\citep{de2019commitmentbank}, RTE~\citep{wang2019superglue}; (2) sentence completion: COPA~\citep{roemmele2011choice}, HellaSwag~\citep{zellers-etal-2019-hellaswag}, Story Cloze~\citep{mostafazadeh-etal-2016-corpus}; (3) coreference resolution: WSC, Winogrande~\citep{levesque2012winograd}; and (4) word sense disambiguation: WIC~\citep{pilehvar-camacho-collados-2019-wic}. We access them using Hugging Face Datasets~\citep{lhoest-etal-2021-datasets} and most of them are from the SuperGLUE benchmark~\citep{wang2019superglue}. All of these datasets are classification-based, predicting a discrete label from a finite set. Each of these datasets has a diverse set of prompts provided by the Public Pool of Prompts~\citep{sanh2021multitask}
The number of prompts ranges from 4 to 15. 
Please refer to Appendix~\ref{app:data} for detailed statistics of these datasets.

\paragraph{Setup:}
We build our method on top of the PLM T0~\citep{sanh2021multitask}. T0 is an adapted version of the pretrained T5 model~\citep{raffel2020exploring} that is continually trained on multiple tasks with supervised, prompt-formatted examples. T0 outperforms GPT3~\citep{brown2020language} and demonstrates state-of-the-art performance in zero-shot task generalization. All the tasks that we are studying are not included in T0's training data. We focus our major study on the T0 model version with 3 billion parameters (T0-3B), while we also include results using the largest T0 model with 11 billion parameters (T0-11B) on some datasets, due to the high computational cost of training T0-11B.  
The hyperparameters (e.g.~the optimization hyperparameters) are tuned on the RTE dataset with its validation set and fixed for all other datasets. 
We use a bottleneck dimension of 1 for LoRA. Complete setup details can be found in Appendix~\ref{app:setup}.

\subsection{Evaluation}
\paragraph{Metrics:}
We use accuracy as the metric for all datasets. We report two different types of accuracy given that we have multiple prompts. The \emph{ensemble accuracy} (Ens.) averages the output distributions of multiple prompts and makes predictions according to it. 
Ensembling multiple prompts has been explored before and found superior to using a single prompt~\citep{jiang-etal-2020-know,qin-eisner-2021-learning}. 
The \emph{median accuracy} (Med.) within the set of prompts  serves as a proxy for the expected performance when users specify a single prompt and input a prompt-formatted example. 
As our approach assumes availability of a set of prompts for the downstream task, and it is relatively cheap to craft several prompts for a task, ensemble prediction is the better option given input $\rvx$, and it does empirically yield higher accuracy overall than the median for both the baseline and our method. 
Therefore, we will report both numbers but mainly discuss ensemble accuracy. 
We report these metrics on the validation split of each dataset. We run the experiments with 3 random seeds and report the mean and standard deviation.

\paragraph{Evaluation scenarios:}
We provide our methods with different unlabeled sources which lead to two practical scenarios during evaluation: (1) \emph{training-time tuning}: we use the unlabeled training split from the corresponding dataset to train the model. This is similar to traditional settings where training and test data are different; and (2) \emph{test-time tuning}~\citep{sun2020test,wang2021tent}: we directly adapt the PLM on the test data. This setting is reasonable, as we will always have access to the test inputs at test time. Intuitively, the unlabeled test sample $\rvx$ often provides hints about the distribution it was drawn, suggesting that we may update the model before making the prediction. This scenario is attractive since it alleviates the common distribution mismatch issue when there is a distribution shift between the training and test data. 
Compared to training-time tuning, test-time tuning typically uses less unlabeled data in our experiments since it uses the validation split itself.
In the major experiments, we focus on the offline test-time tuning where we assume access to the entire test data\footnote{To clarify, test data is not the test split of the dataset, but the data that we evaluate on, i.e.~the validation split.} and train our approach on all test examples, while in \textsection\ref{sec:analysis} we will discuss the potential for online adaptation where data arrives in a stream. 
\paragraph{Baselines:}
As far as we know, there is no prior work studying unsupervised approaches for this prompt-based task generalization setting, thus T0 is the main baseline that we compare our approach against. However, we still implement an ablation baseline, self distillation, to separate the improvement from optimizing prompt consistency and pseudo-label distillation. Specifically, self distillation minimizes the same loss as in Eq.~\ref{eq:pairwise-distill} but with $r^{(i)}=r^{(j)}$ -- instead of pairwise distillation, the prompt always distills its own prediction to itself. This baseline can be viewed as a prompt version of self-training, which has proven to effectively utilize unlabeled data~\citep{He2020Revisiting,xie2020self,zhang2020pushing}. We report self distillation results in the training-time tuning setting only for simplicity. 

\subsection{Results}

\begin{table}[!t]
    \centering
\resizebox{0.9 \columnwidth}{!}{
\begin{tabular}{lcccc}
\toprule
          & \multicolumn{2}{c}{T0-11B}                          & \multicolumn{2}{c}{Swarm Dist.}           \\
Dataset   & \multicolumn{1}{c}{Ens.} & \multicolumn{1}{c}{Med.} & \multicolumn{1}{c}{Ens.} & \multicolumn{1}{c}{Med.} \\
\midrule
WSC       & 63.5                    & 62.5                     & 65.4\up{1.9}                     & 62.0\down{0.5}                     \\
RTE       & 83.8                    & 82.0                    & 86.6\up{2.8}                     & 85.0\up{3.0}                     \\
HellaSwag & 34.4                     & 33.6                     & 45.0\up{10.6}                     & 43.0\up{9.4}                     \\
WIC       & 57.2                     & 56.8                     & 62.1\up{4.9}                     & 60.7\up{3.9}   \\        
\bottomrule  
\end{tabular}}
    \caption{\label{tab:11b}Accuracy based on T0-11B.}
\end{table}

\begin{table}[!t]
    \centering
    \resizebox{1 \columnwidth}{!}{
    \begin{tabular}{lrrrr}
        \toprule
            & T0-3B & No Shift & SD on MNLI & SD on QNLI \\
        \midrule
    RTE     & 64.6  & 75.2     & 75.5            & 72.9            \\
    ANLI R1 & 34.6  & 38.4     & 38.2            & 37.4            \\
    ANLI R2 & 33.7  & 37.9     & 38.5            & 35.3            \\
    ANLI R3 & 34.7  & 34.0     & 36.9            & 34.4    \\
    \bottomrule       
    \end{tabular}}
    \caption{\label{tab:shift} Ensemble accuracy on the distribution shift setting based on T0-3B. ``No Shift'' represents the original setting where we train the swarm distillation loss on the training split from the same dataset as the test examples. ``SD on MNLI/QNLI'' represents swarm distillation trained on the training split of MNLI/QNLI. }
\end{table}

\begin{table*}[!t]
    \centering
    \resizebox{1 \textwidth}{!}{
\begin{tabular}{lrrrrrrrrrrrr}
\toprule
                   & RTE   & CB & ANLI R1 & ANLI R2 & ANLI R3 & COPA  & HS & Story. & Wino. & WSC   & WIC   & Avg.   \\
\midrule
T0-3B              & 0.644 &  0.440  & 0.221   & 0.189   & 0.170    & 0.586 & 0.164     & 0.765       & 0.396 & 0.255 & 0.398 & 0.384 \\
Swarm Dist. & 0.662 &  0.254  & 0.145   & 0.156   & 0.177   & 0.699 & 0.402     & 0.862       & 0.509 & 0.462 & 0.517 & {\bf 0.440} \\
\bottomrule
\end{tabular}}
    \caption{\label{tab:consistency}Fleiss' kappa on 11 datasets based on T0-3B. Swarm distillation is trained on the training split of the respective dataset.}
\end{table*}

\paragraph{How well does swarm distillation work?} 
We first compare swarm distillation against the T0-3B baseline. 
As shown in Table~\ref{tab:res-main}, the ensemble accuracy of swarm distillation exceeds the T0-3B baseline on 9 out of 11 datasets in both training- and test-time tuning settings. Particularly, our approach improves the zero-shot performance on RTE by around 10 absolute points in all cases. Our approach slightly hurts ensemble accuracy of ANLI R3 and median accuracy of CB, but is overall comparable on these two datasets.
Compared to self distillation, swarm distillation outperforms it on 9 out of 11 datasets in terms of ensemble accuracy, by up to 10.3 absolute points. These results further confirm the effectiveness of encouraging prompt consistency.
We note that swarm distillation severely fails on WSC with a 10-point accuracy decrease compared to both T0 and self distillation, this is because Fleiss' kappa selects a bad model checkpoint, while our approach actually improves the performance on WSC in the middle of training as we will discuss more in \textsection\ref{sec:analysis}.
Notably, our approach is helpful on several datasets where T0-3B only shows nearly chance accuracy, such as ANLI R1/R2/R3 (3 labels), HellaSwag (4 labels), Winogrande (2 labels), and WIC (2 labels). 
In addition, we observe that swarm distillation in the test-time tuning setting performs comparably well to the training-time one despite using much less training data, as shown in Appendix~\ref{app:data}. 
It is worth noting that prompt-based zero-shot task generalization is challenging, for example, T0 with even 11 billion parameters reports a median accuracy of only $\sim40$ on ANLI R1/R2/R3, 33.7 on HellaSwag, and 57.2 on WIC~\citep{sanh2021multitask}. Our results are surely still far from satisfactory, yet we hope to inspire future research to explore unlabeled data to build better zero-shot learners.    

\paragraph{Scaling to 11B parameters:}
We now evaluate our method based on the largest version of T0 model, T0-11B. T0-11B is a very powerful zero-shot baseline that greatly outperforms GPT3. Due to the expensive computation to train T0-11B, we use one dataset per task, a total of 4 datasets as our benchmark, and only run with one random seed in the test-time tuning setting. Results are shown in Table~\ref{tab:11b}. Swarm distillation outperforms T0-11B on all 4 datasets in terms of ensemble accuracy, and notably, improves the ensemble accuracy on HellaSwag from 34.4 to 45.0 without any annotation. Table~\ref{tab:res-main} and Table~\ref{tab:11b} demonstrate the effectiveness of swarm distillation across different model sizes.

\paragraph{Robustness of swarm distillation:}
The main experiments so far train and test on examples from the same dataset, thus the results are based on the assumption that the training and test distributions are similar. 
To relax this assumption, here we evaluate swarm distillation with a more difficult setting by collecting unlabeled training examples from other datasets that are not listed in \textsection\ref{sec:setup}. 
Specifically, we focus on the NLI task, train swarm distillation on the unlabeled training split of MNLI~\citep{N18-1101} and QNLI~\citep{wang2018glue} respectively, and evaluate it on RTE and ANLI.
Table~\ref{tab:shift} shows that our proposed method still performs well in such a distribution shift setup –- the models trained on MNLI or QNLI help improve the T0-3B baseline in most cases. The results from MNLI training are generally comparable to the original ``No Shift'' numbers, while QNLI training causes a mild accuracy drop.

\begin{figure*}[!t]
    \centering
    \includegraphics[width=0.9\textwidth]{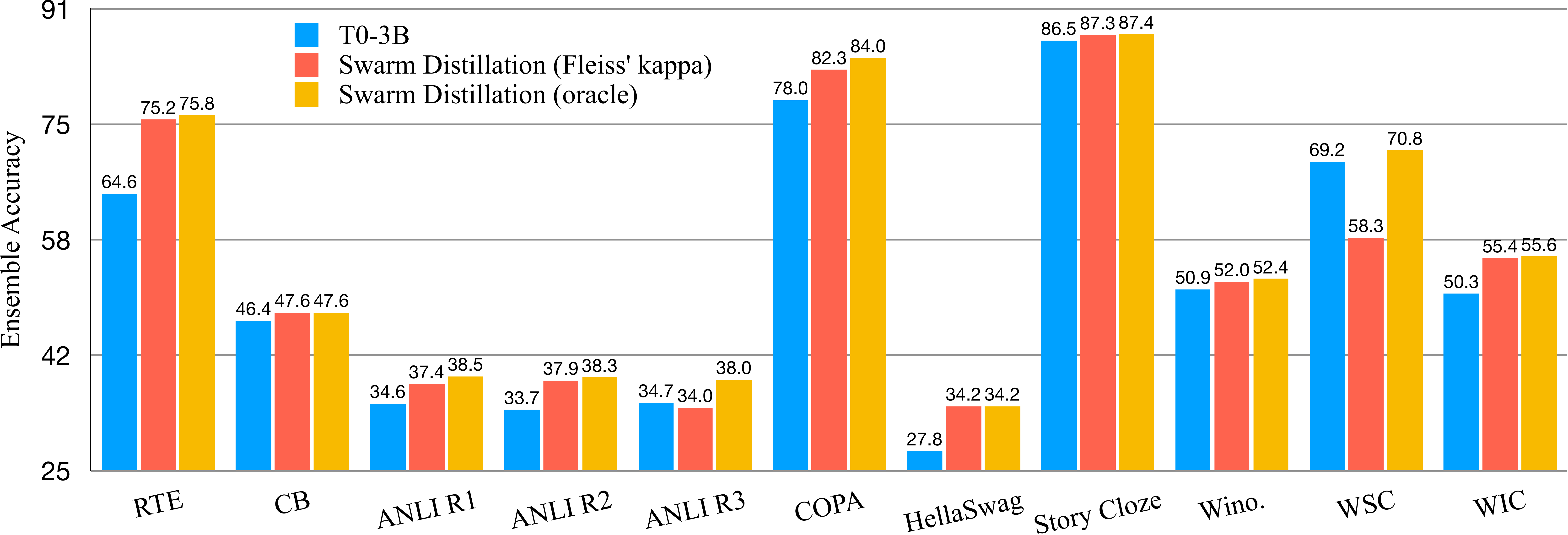}
    \caption{Analysis results to compare the model checkpoints selected by the unsupervised criterion Fleiss' kappa with the oracle model checkpoints selected by validation accuracy.
    }
    \label{fig:res-criterion}
\end{figure*}

\subsection{Analysis}
\label{sec:analysis}
\paragraph{Are predictions more consistent across prompts after swarm distillation?}
We wonder whether the gains of swarm distillation are attained together with more consistent predictions across different prompts. To this end, we report Fleiss' kappa, a commonly used metric for group agreement as detailed in \textsection\ref{sec:train}. Results are shown in Table~\ref{tab:consistency}. Fleiss' kappa on 8 out of 11 datasets increases after swarm distillation, which boosts the averaged Fleiss' kappa of T0-3B by 14.6\% relatively. This implies that swarm distillation facilitates prompt consistency, and potentially improves the robustness of PLMs to different wording of prompts.

\paragraph{Does the unsupervised criterion select the best model checkpoint?}
In \textsection\ref{sec:train}, we discussed using Fleiss' kappa to select the best model checkpoint for evaluation, here we report the oracle accuracy numbers obtained by selecting the model checkpoint with the best validation accuracy, and compare it to the one selected by Fleiss' kappa.
We compare the ensemble accuracy using T0-3B in the training-time tuning setting, with results in Figure~\ref{fig:res-criterion}. On most of the datasets, Fleiss' kappa is able to achieve numbers close to the best ones. On all 11 datasets, our oracle number outperforms the T0-3B baseline. In Table~\ref{tab:res-main} we show that swarm distillation hurts the performance on WSC a lot, while in Figure~\ref{fig:res-criterion} swarm distillation (oracle) in fact outperforms T0-3B, implying that the issue lies on model selection. 
Therefore, swarm distillation could potentially work better if an annotated dev set is available or when it is combined with other techniques in few-shot learning settings, where good checkpoints may be selected out more easily. 


\begin{figure}[!t]
    \centering
    \begin{subfigure}[b]{0.47\columnwidth}
        \includegraphics[width=1\textwidth]{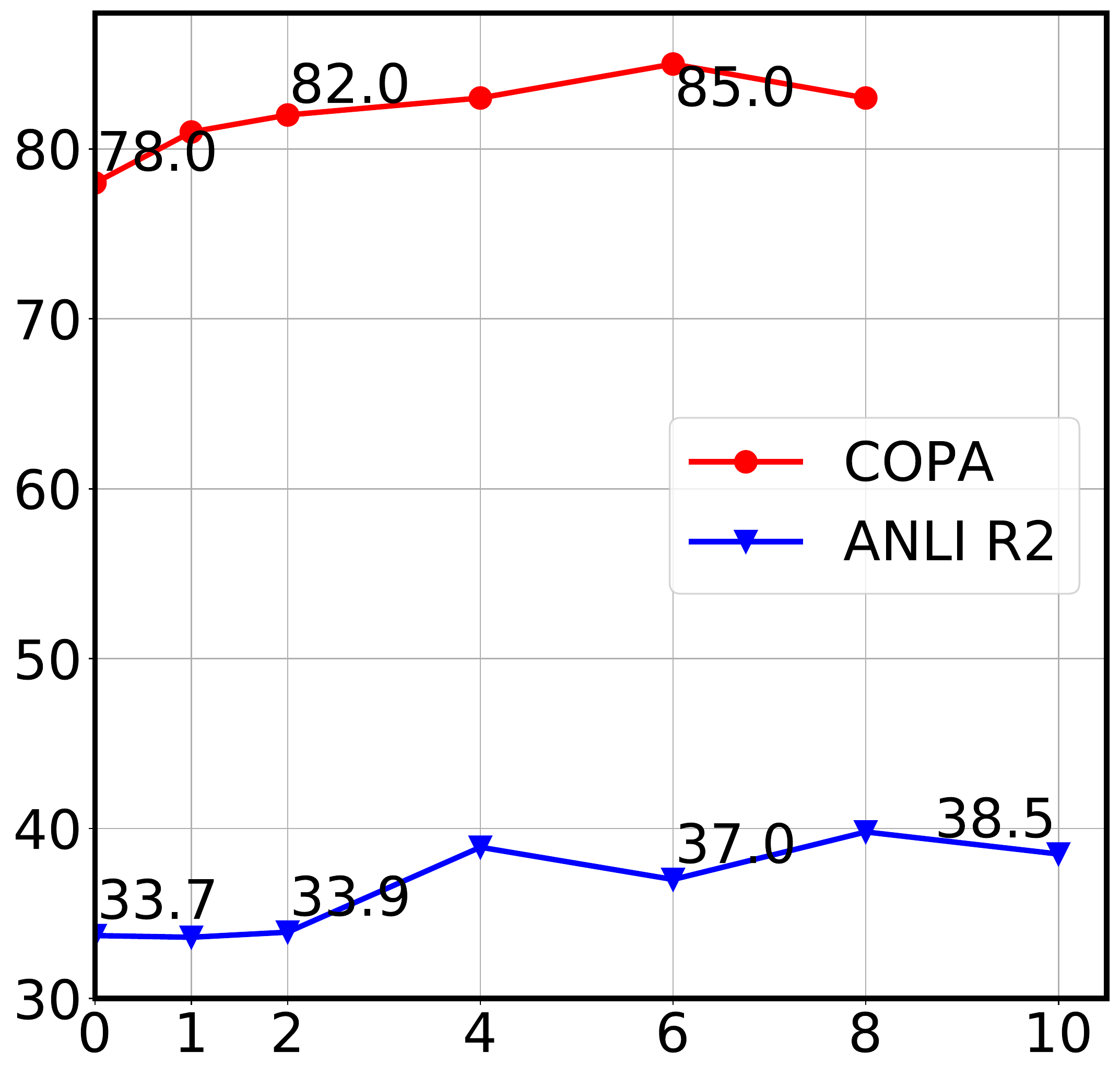}
        \caption{Accuracy v.s. \#prompts} 
        \label{fig:res-prompt}
    \end{subfigure}
    \hfill
    \begin{subfigure}[b]{0.47\columnwidth}
        \includegraphics[width=1\textwidth]{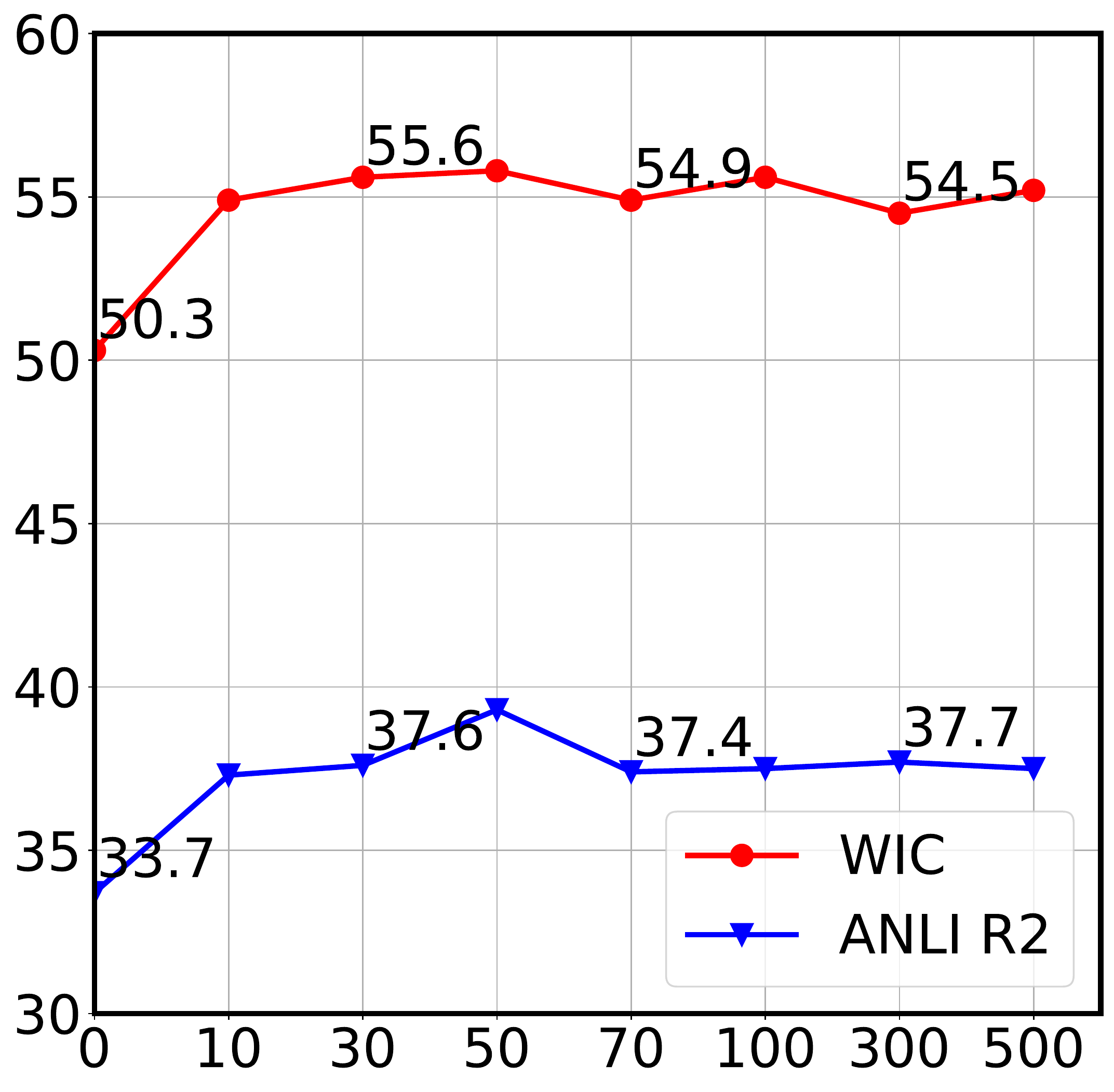}
        \caption{Accuracy v.s. \#examples} 
        \label{fig:res-example}
    \end{subfigure}
    \label{fig:res-analysis}
    \caption{Ensemble accuracy of swarm distillation on three example datasets based on T0-3B, demonstrating the effect of prompt size and unlabeled data size.
    }
\end{figure}

\paragraph{How many prompts do we need?}
Our approach requires a diverse set of prompts to regularize prompt consistency. Here we perform ablation experiments to understand the effect of the number of prompts on the performance. We take COPA and ANLI R2 as example datasets which have 8 and 15 prompts, respectively. We then vary the number of available prompts by randomly sampling a subset of prompts before training. 
We report the ensemble accuracy of swarm distillation (train) in Figure~\ref{fig:res-prompt}. On both COPA and ANLI R2, we observe gains as we increase the number of prompts from 0 (the baseline), yet the performance saturates very quickly and relatively stabilizes when we provide 4 prompts. This implies that swarm distillation is not prompt-hungry and could work well with a small number of prompts. 
We note the with one prompt here Eq.~\ref{eq:pairwise-distill} degenerates to a weaker version of self distillation compared to the one in Table~\ref{tab:res-main} -- self distillation in Table~\ref{tab:res-main} utilizes all prompts during training while we assume access to only one prompt here. 

\paragraph{How many unlabeled examples do we need?}
We measure the effect of unlabeled data size. Specifically, we randomly sample a subset of examples from the train split for training and report results on the entire validation dataset.
Results on WIC and ANLI R2 are shown in Figure~\ref{fig:res-example}. Notably, swarm distillation is able to outperform the baselines (\#examples=0) by a large margin on both datasets with only 10 unlabeled examples, and the performance starts to saturate quickly afterward. These results suggest that swarm distillation is not data-hungry and works reasonably well with few \emph{unlabeled} examples, allowing swarm distillation to remain as a relatively light approach while typical unsupervised training (e.g. pretraining) often requires a large amount of data and computation. 
Also, we argue that the phenomenon demonstrated in the results implies that swarm distillation may be applied to the online setting of test-time tuning, where the batches of test data arrive in a stream. Online test-time tuning is a practical setting in real life, and we leave it as future work to study. 

\section{Discussion}
\label{sec:discuss}
In this paper, we explore prompt consistency regularization to make PLMs better zero-shot learners.
Our approach, swarm distillation, utilizes unlabeled examples to attain zero-shot gains. While we use swarm distillation in a post-adaptation manner, it could be potentially combined with the pretraining objectives in the pretraining stage (e.g. the multi-prompt training loss~\citep{sanh2021multitask,wei2021finetuned}), or even with annotated data in few-shot learning settings. Combining swarm distillation with these other losses may easily bypass the model collapse issue since the other loss typically discourages the collapsed local optimum. 

\section*{Acknowledgement}
We thank Victor Sanh and Colin Raffel for help on using the T0 model. This work was supported in part by the CMU-Portugal MAIA Project, a Baidu PhD Fellowship for Junxian He, and a CMU Presidential Fellowship for Chunting Zhou.

\section*{Limitations}
There are two limitations of our work: (1) Because our method is operated in a fully unsupervised manner, there is no supervised development data for us to either select the best model or tune hyperparameters. Thus, we propose to use Fleiss' Kappa as our unsupervised development metric for model selection, which attains decent performance in most cases. However, we also observe on very few datasets that the proposed metric fails to select the best checkpoints and hurt the model's performance. As discussed in~\S\ref{sec:analysis}, our method can be combined with few-shot learning where a few labeled data are provided, and we believe this can largely alleviate the issues of model selection in the unsupervised setting.
(2) The other limitation and at the same time an advantage of our method is that the proposed method can work well even with 10 unlabeled data points. This certainly makes our method a good candidate for the online setting where batches of test data come in a stream. However, as we discussed in~\S\ref{sec:analysis}, the performance of our model saturates quickly as we increase the number of unlabeled data, which means the performance of our method cannot scale well with tons of unlabeled data like self-supervised pretraining.
As discussed in \S\ref{sec:discuss}, we expect combining our method with few-shot learning setting / pre-training can lead to further improvements as the supervised signals may guide the model to a better local optimum.

\section*{Ethics Statement}
Similar to T0, this work aims to produce an open-ended system that could perform all text-based tasks through designing different prompts. While the performance of GPT3, T0, and this work is far from the practical level on unseen tasks, we expect a greatly improved prompt-based system in the future could be built to help perform many daily tasks in real life.
However, the resulted model in this paper also admits the same ethics concerns that T0 has. For example, the unrestricted use of prompts may easily trigger offensive generations or private information leakage, and how to fix unwanted LM behaviors is still an active research problem~\citep{liu-etal-2021-dexperts,perez2022red}.

\bibliography{anthology,custom}
\bibliographystyle{acl_natbib}

\clearpage
\newpage
\appendix
\begin{table}[!t]
    \centering
    \resizebox{0.5 \textwidth}{!}{
    \begin{tabular}{llcccc}
    \toprule
     Task & Dataset  & \#train set & \#validation set & \#labels & \#prompts                                           \\
    \midrule
    \multirow{5}{*}{NLI} & RTE & 2,490 & 277 & 2 & 10 \\
     & CB          & 250 & 57 & 3 & 15 \\ 
     & ANLI R1     & 16,946 & 1000 & 3 & 15 \\
     & ANLI R2     & 45,460 & 1000 & 3 & 15 \\ 
      & ANLI R3     & 100,459 & 1200 & 3 & 15 \\
    \midrule
    \multirow{3}{*}{Compl.}                  & COPA        & 400 & 100 & 2 & 8 \\
                                             & HellaSwag   & 39,905 & 10,042 & 4 & 4 \\
                                             & Story Cloze & - & 1,871 & 2 & 5\\
    \midrule
    \multirow{2}{*}{Coref.}                  & Winogrande & 40,398 & 1,267 & 2 & 5 \\
                                             & WSC  & 554 & 104 & 2 & 10 \\
    \midrule
    WSD                                      & WIC &  5,428 & 637 & 2 & 10\\
    \bottomrule 
    \end{tabular}}
    \caption{\label{tab:data}Statistics of the datasets}
\end{table}

\begin{table*}[!t]
    \centering
\resizebox{1 \textwidth}{!}{
\begin{tabular}{lrrrrrrrrrrr}
\toprule
                 & RTE  & CB   & ANLI R1 & ANLI R2 & ANLI R3 & COPA & HS & Story. & Wino. & WSC  & WIC  \\
\midrule
T0-3B & 64.6 & 46.4 & 34.6    & 33.7    & 34.7    & 78.0   & 27.8      & 86.5        & 50.9  & 69.2 & 50.3 \\
\midrule
Full fine-tuning & 59.6 & 8.9 & 33.3    & 33.5    & 33.3    & 54.0 & 25.7      & 51.5        & 50.4  & 36.5 & 50.0\\
\ + LoRA             & 54.0 & 44.6 & 33.3    & 33.3    & 34.4    & 82.7 & 33.6      & 87.4        &  52.0 & 36.5 & 50.0 \\
\ + model selection & 75.8 & 35.7 & 37.0 & 33.5 & 33.3 & 80.0 & 32.2 & 86.5 & 50.8 & 71.2 & 54.6 \\
\ + LoRA + model selection & 75.2 & 47.6 & 38.4 & 37.9 & 34.0 & 82.7 & 34.2 & 87.3 & 52.0 & 58.3 & 55.4\\ 
\bottomrule
\end{tabular}}
\caption{\label{tab:lora-ablation}Ablation results on LoRA and unsupervised model selection. The training objective is the swarm distillation loss. Numbers are ensemble accuracy in the training-time tuning setting based on T0-3B. ``Full fine-tuning'' updates all the model parameters, while ``+LoRA'' means that we freeze the T0 parameters and only update the LoRA parameters.}
\end{table*}

\section{Datasets}
\label{app:data}
We present the statistics of the 11 datasets in Table~\ref{tab:data}.
For the training-time tuning scenario, we use up to 10,000 data points from the training set for training if the train set contains more than 10,000 data points.


\section{Experimental Setup}
\label{app:setup}
\subsection{LoRA Setup}
We use LoRA~\citep{hu2021lora} as our parameter-efficient tuning model and set the bottleneck dimension of LoRA weight matrices to be 1 for both 3B and 11B models. 
We emphasize that the linear mapping matrix $B$ (or $A$) in LoRA needs to be initialized as a zero matrix to ensure the output distribution after adding LoRA layers is the same as the original PLM before training, otherwise, the zero-shot ability of PLMs would be broken upon initialization and there is no supervision to learn it back.
For both models, we set the dropout probability for the the LoRA intermediate representations to be 0.3.
Let $\alpha$ denote the scaling factor of LoRA that is used to scale the output of the LoRA layer before adding to the hidden states of the pre-trained model. 
We set $\alpha$ to be 4 and 2 respectively for the 3B and 11B model.
The peak learning rates of the 3B and 11B models are set to be 3e-5 and 5e-5 respectively with a warm-up stage of 100 steps and polynomial learning rate scheduler.
We train for a maximum of 1,500 steps. 
Note that the hyperparameters for the 3B model is tuned on the RTE dataset and used for other datasets. We did not tune the hyperparameters of the 11B model.

\subsection{Implementation Details}
The reported T0 baseline numbers are obtained from our own running using the released T0 weights. 
We are able to reproduce the numbers reported in~\citet{sanh2021multitask}, except for COPA where our T0 median number is higher than the originally reported one.

During training, at each update we first sample one input example $\rvx$ and apply all the prompts to reformat it as $r_x^{1}(\rvx), \cdots, r_x^{K}(\rvx)$, 
then we perform inference for them and randomly shuffle the predictions.  
Next we iterate over them with a batch size of 5/10 (3B/11B)\footnote{Because the GPU memory sometimes cannot handle all the prompts within one batch.} and use the shuffled predictions to supervise them to compute the distillation loss, this implements the swarm distillation mechanism in Eq.~\ref{eq:pairwise-distill} and amounts to approximating the expectation over paired prompts with $K$ random pairs. 
We accumulate the gradients for 16 steps for one update so that each gradient descent is computed from 16 data examples. 
And we use 1 A40 GPU (45GB memory) to train the 3B model and 4 A40 GPUs with DeepSpeed Zero-2~\citep{ren2021zero} to train the 11B model.
In general, training converges pretty fast and takes around 1 - 3 GPU hours for the 3B model and 2 - 6 hours for the 11B model depending on early stop points of different datasets. 
We use Adam~\citep{kingma2014adam} as the optimizer with $\beta_1=0.9$, $\beta_2=0.98$ and $\epsilon=1e-6$.

For the Transformer~\citep{vaswani2017attention} models with model dimension $d$, the feed-forward intermediate dimension $m$ and number of layers $l$, the additional parameters used in LoRA with bottleneck dimension $b$ is calculated as $b*(m+d)*2*l*2$. As we set $b$ to be 1 for both the 3B and 11B models, the additional number of LoRA parameters is 1,671,168 for the T0-3B model ($d=1024, m=16384, l=24$) and 6,389,760 for the T0-11B model ($d=1024, m=65536, l=24$).

\section{Ablation on LoRA and Model Selection}
\label{app:lora-full}
We report the ablation results on LoRA and unsupervised model selection in Table~\ref{tab:lora-ablation}.
Full fine-tuning hurts the T0-3B performance on all datasets --  actually it collapses on almost all the datasets when we check its predictions, which could partially explain the low accuracies. 
Using LoRA alone is able to improve full fine-tuning generally and outperforms the T0-3B baseline sometimes. 
Moreoever, we find that unsupervised model selection is very effective to mitigate collapse and greatly improves full fine-tuning results. 
Finally, combining LoRA and unsupervised model selection gives the best results overall on these 11 datasets.

We clarify that the results in Table~\ref{tab:lora-ablation} are only for analysis purpose to better understand the effect of different components of our method, but were not used by us to make design decisions in our preliminary experiments-- as stated in \textsection\ref{sec:train}, we use LoRA because it collapses less often than full fine-tuning and develop a unsupervised model selection criterion since LoRA still collapses on some datasets. To judge collapse or not, we simply checked the model predictions, to see if the predictions for all the examples are almost the same.

\end{document}